\documentclass[letterpaper, 10 pt, conference]{ieeeconf}  

\IEEEoverridecommandlockouts                          

\usepackage{graphics} 
\usepackage{epsfig} 
\usepackage{mathptmx} 
\usepackage{times} 
\usepackage{amsmath} 
\usepackage{amssymb}  
\usepackage{bm}
\usepackage{dsfont}
\usepackage{xcolor}
\usepackage{hyperref}

\title{\LARGE \bf
GRANP: A Graph Recurrent Attentive Neural Process Model for Vehicle Trajectory Prediction
}

\author{Yuhao Luo$^{1}$, Kehua Chen$^{2,3}$,~\IEEEmembership{Graduate Student Member,~IEEE}, 
Meixin Zhu$^{1,4*},~\IEEEmembership{Member,~IEEE}$
\thanks{(Corresponding author: Meixin Zhu)}
\thanks{$^{1}$Yuhao Luo and Meixin Zhu are with Intelligent Transportation Thrust, Systems Hub, The Hong Kong University of Science and Technology (Guangzhou), Guangzhou, China.}%
\thanks{$^{2}$Kehua Chen is with Division of Emerging Interdisciplinary Areas (EMIA), Interdisciplinary Programs Office, The Hong Kong University of Science and Technology, Hong Kong, China.}%
\thanks{$^{3}$Kehua Chen is also with Department of Civil and Environmental Engineering, The Hong Kong University of Science and Technology, Hong Kong, China.}%
\thanks{$^{4}$Meixin Zhu is also with Guangdong Provincial Key Lab of Integrated Communication, Sensing and Computation for Ubiquitous Internet of Things, Guangzhou, China.}
}

\begin{document}

\maketitle
\thispagestyle{empty}
\pagestyle{empty}

\begin{abstract}
As a vital component in autonomous driving, accurate trajectory prediction effectively prevents traffic accidents and improves driving efficiency. To capture complex spatial-temporal dynamics and social interactions, recent studies developed models based on advanced deep-learning methods. On the other hand, recent studies have explored the use of deep generative models to further account for trajectory uncertainties. However, the current approaches demonstrating indeterminacy involve inefficient and time-consuming practices such as sampling from trained models. To fill this gap, we proposed a novel model named Graph Recurrent Attentive Neural Process (GRANP) for vehicle trajectory prediction while efficiently quantifying prediction uncertainty. In particular, GRANP contains an encoder with deterministic and latent paths, and a decoder for prediction. The encoder, including stacked Graph Attention Networks, LSTM and 1D convolutional layers, is employed to extract spatial-temporal relationships. The decoder is used to learn a latent distribution and thus quantify prediction uncertainty. To reveal the effectiveness of our model, we evaluate the performance of GRANP on the highD dataset. Extensive experiments show that GRANP achieves state-of-the-art results and can efficiently quantify uncertainties. Additionally, we undertake an intuitive case study that showcases the interpretability of the proposed approach. The code is available at \url{https://github.com/joy-driven/GRANP}.

\end{abstract}

\section{INTRODUCTION}
In autonomous driving, trajectory prediction aims to predict the future states of the surrounding vehicles to alleviate driving risks and improve travel efficiency. As a vital component, researchers have endeavored to achieve accurate and efficient trajectory prediction recently. Early studies primarily used physics-based models to predict trajectories, such as Kalman Filtering \cite{ammoun2009real}. However, physics-based models have low complexity, low accuracy, and cannot capture complex spatial-temporal features. With the development of Artificial Intelligence, recent studies have focused on data-driven approaches. For instance, \cite{park2018sequence} used a seq2seq model based on Long Short-Term Memory (LSTM) for prediction. \cite{li2019grip} applied a Graph Convolutional Network to learn the spatial features and utilized a LSTM decoder to generate trajectory. However, most deep-learning models fail to quantify prediction uncertainty. The ability to quantify uncertainty is significant, especially in complex driving scenarios. If deterministic models cannot provide accurate predictions, the misinformation may lead to incorrect decisions or even collisions. To fill this gap, deep generative models have been applied to encompass uncertainties. For example, \cite{choi2021dsa} applied a Convolutional Neural Network (CNN) to identify the driving style and then used a Conditional Generative Adversarial Network (CGAN) to generate the predicted trajectories from the distribution. \cite{de2022vehicles} utilized the Variational Auto Encoder (VAE) to encode the behavior of traffic participants and then predict trajectories via a decoder. However, most generative models learn implicit distributions through real data, and quantifying uncertainties often relies on sampling, which can be extremely inefficient.


To overcome the limitations of existing methods, we proposed a novel method GRANP, consisting of Recurrent Attentive Neural Process (RANP), Graph Attention Networks (GAT), 1D convolutional layers and Long-Short Term Memory (LSTM). Among them, RANP is the core structure of GRANP, composed of an encoder with deterministic and latent paths. The deterministic path generates a deterministic representation for contextual information, and the latent path learns a latent distribution, enabling GRANP to quantify uncertainty. Additionally, we utilize GAT to capture the social interactions among traffic participants. 

To sum up, the main contributions of our work are summarized below:
\begin{itemize}
    \item We propose a Graph Recurrent Attentive Neural Process model. GRANP has the competitive advantage of NPs to directly quantify and visualize prediction uncertainty with robust and stable performance across a wide range of scenarios. 

    \item We utilize GAT, LSTM and triple 1D convolutional layers to efficiently capture the spatial-temporal relationships.

    \item We evaluate the performance of our model on the highD dataset and compare it with several state-of-the-art models, showing that GRANP outperforms the baseline models. Additionally, we quantify the uncertainty of the prediction, demonstrating the strengths of our model.
\end{itemize}

\section{RELATED WORK}
Researchers have proposed various methods to achieve trajectory prediction and these methods can be divided into 4 types, including physics, machine learning (ML), deep learning (DL), reinforcement learning (RL). 
In this section, we provide a brief overview of studies pertaining to these methods.

\subsection{Physics-based Models}
Physics-based methods were popular for trajectory prediction in early studies. The physics-based models involve dynamics models and kinematics models, including linear Kalman filter \cite{ammoun2009real}, motion models \cite{schubert2008comparison} and Monte Carlo models \cite{broadhurst2005monte}. However, the prediction accuracy of these models is relatively low and most of them are only well-suited for short-term forecasting.

\subsection{Machine Learning-based Models}
With the explosion of data-driven models, researchers leveraged various machine learning methods for trajectory prediction. \cite{hewing2020simulation} proposed Gaussian Process (GP)-based models accounting for correlations between the used evaluations. \cite{deo2018would} used Hidden Markov Model (HMM) and variational Gaussian mixture models (VGMM) to predict trajectory. Compared with physics-based Models, machine learning-based models take more factors into account, successfully improving trajectory prediction accuracy.

\subsection{Deep Learning-based Models}
As the most popular method, deep learning methods are widely used in trajectory prediction. \cite{zyner2018recurrent} employed stacked Long Short-Term Memory (LSTM) with a single full connected layer for prediction. Considering the influence of driving style, \cite{xing2019personalized} used Gaussian Mixture Model (GMM) before predicting trajectory to classify the drivers' driving style. Moreover, many studies have been devoted to combining CNN and Recurrent Neural Network (RNN). \cite{deo2018convolutional} employed LSTM encoder to extract the temporal information of surrounding agents and then used CNNs to learn the spatial information. Additionally, other techniques, such as Graph Neural Network (GNN) and Generative Model (GM), have been applied in this field. Due to their powerful expressiveness, deep learning-based models significantly outperform physics-based models and machine learning-based models. 


\section{METHODOLOGY}
\begin{figure*}[t]
  \centering
  \includegraphics[width=0.9\textwidth]{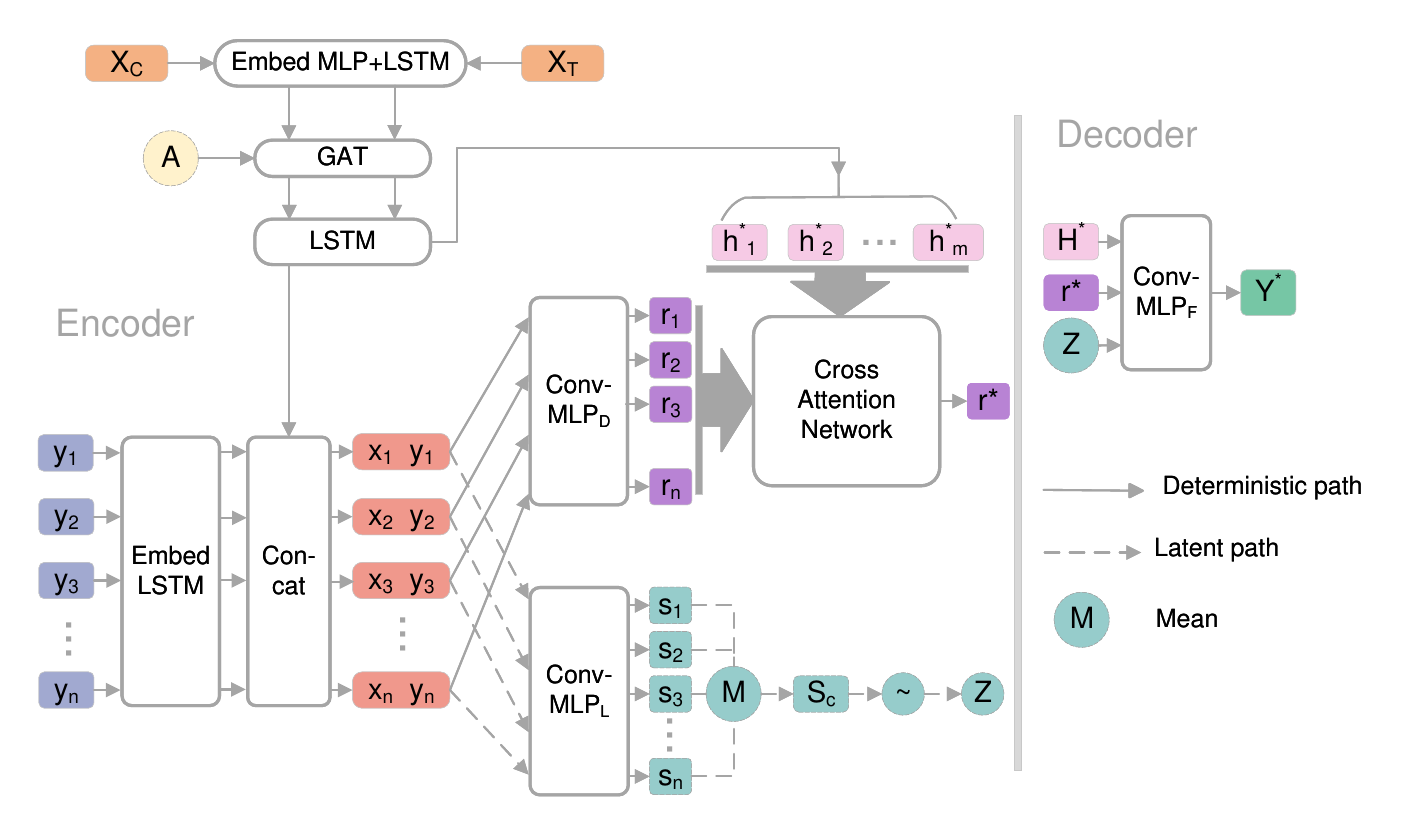}
  \caption{Overall framework of GRANP. GRANP primarily follows the structure of RANPs, including an encoder and a decoder with deterministic and latent paths. More detail, the Conv-MLP module contains a triple 1D convolutional layers and 4 full connected layers, and GAT module consists of two graph attention layers.}
  \label{fig:model}
\end{figure*}

In this section, we first define the trajectory prediction problem. Afterwards, we provide details related to the proposed GRANP as Fig. \ref{fig:model} shows. In detail, GRANP includes following modules: an encoder, consisting of latent and deterministic paths and a decoder.

The preliminary steps for both the deterministic path and the latent path are the same. First, both historical and future trajectories in the training set are transformed into contextual pairs through the embedding layer, and we use MLP to resolve the length inconsistency. The GAT and LSTM are applied to capture the spatial-temporal features. Moreover, the deterministic path employs triple 1D convolutional layers and MLP to encode each contextual pair. The latent path owns a similar structure to the deterministic path, and finally generates a latent distribution to indicate uncertainties. Given target historical trajectories, we integrate the contextual embeddings and target embeddings with the Cross Attention Network, and generate a deterministic embedding. Then, we input the deterministic embedding, target embedding, and sampled latent embedding to the decoder for prediction. Finally, the decoder outputs future trajectories as a Gaussian distribution.

\subsection{Problem Definition}

The vehicle trajectory prediction problem aims to predict their future states based on the past states of traffic participants. In this work, vehicles' historical states are formulated as:
\begin{align}
    X = \left\{ S^1, S^2, \ldots, S^{T_N} \right\} 
\end{align}
where \( X \) denotes the set of historical states, \(T_N \) represents the length of historical time. \( S^t \) (\( t \in \{1, 2, \ldots, T_N\} \)) represents the states of all considered road users at timestamp t, formulated as:
\begin{align}
    S^ t = \left\{ (x^t_0, y^t_0, s^t_0, a^t_0), (x^t_1, y^t_1, s^t_1, a^t_1), \ldots, (x^t_n, y^t_n, s^t_n, a^t_n) \right\} 
\end{align}
where \((x^t_n, y^t_n, s^t_n, a^t_n)\) denotes the lateral coordinate, longitutional coordinate, speed and acceleration of the $n$-th traffic participant at moment $t$, respectively. Similarly, the future states can be expressed as 
\begin{align}
     Y = \left\{ S^{T_N+1}, S^{T_N+2}, \ldots, S^{T_N+T_F} \right\}
\end{align}
where \(T_F \) represents the length of forecasting time. The goal of trajectory prediction is to find a function $\zeta(X)$ mapping the historical states \(X \) to the future states \(Y \).

\subsection{Recurrent Attentive Neural Process (RANP)}
The RANP combines Recurrent Neural Networks (RNNs) and Attentive Neural Processes (ANPs), possessing both the ability of RNNs to deal with time series data and the advantage of NPs to model uncertainty. RNNs function as an embedding layer, capturing the temporal dependency of the input features. ANPs are the central structure of this model, taking features embedded by RNNs as input and generating a Gaussian distribution. The variance of the distribution represents the uncertainty of results and a high uncertainty indicates possible inaccurate prediction. 


In this study, we represent the historical state using four parameters: lateral coordinate \(x\), longitudinal coordinate \(y\), speed \(s\), and acceleration \(a\). For future state, we focus on the lateral \(x\) and longitudinal \(y\) coordinate. Here, \(X = \left\{ \left(x_{t}^i, y_{t}^i, s_{t}^i, a_{t}^i\right) \right\}^{T_N}_{i=1}\) denotes the historical state up to time \(T_N\), and \(Y = \left\{ \left(x_{t}^j, y_{t}^j\right) \right\}^{T_N+T_F}_{j=T_N}\) represents the projected future trajectories from time \(T_N\) to \(T_N+T_F\). During the data preparation stage, we split the data into a context set \( C \) and a target set \( T \). The context set \( C \) is a collection of pairs \((X_{i}, Y_{i})\) where \( i = 1, 2, \ldots, m \), and the target set \( T \) is a collection of pairs \((X_{j}, Y_{j})\) where \( j = 1, 2, \ldots, k \). In NPs, we assume $Y_T$ is conditioned on $X_T$, $X_C$ and $Y_C$.

During training, we first apply an embedding layer to get the sample embedding $H$ ($H_C$ for context set and $H_T$ for target set). In particular, due to the inconsistent timestamp lengths, we apply an additional MLP to $X_C$ and $X_T$ for interpolation. Then, we gain a deterministic function $r^*_C = \psi(H_C, Y_C, H_T)$ integrating contextual and target information, and a latent variable $z \sim \mathcal{N}(\mu,  \sigma^2)$ by passing through a deterministic path with cross-attention and a latent path respectively. Eventually, the future trajectory $Y_T$ is modeled conditioned on the aforementioned known parameters:
\begin{align}
    p(Y_T | X_T, X_C, Y_C) = \int p(Y_T | H_T, H_C, Y_C, r^*_C, z)q(z | H_C, Y_C) \, dz    
\end{align}
At inference time, our model is optimized using the Evidence Lower Bound (ELBO):
\begin{equation}
\begin{aligned}
    \log p(Y_T | X_T, X_C, Y_C) \geq & \ \mathbb{E}_{q(z|S_T)} \left[ \log p(Y_T | X_T, r^{*}_{C}, z) \right] \\
    & - D_{KL} \left( q(z|S_T) || q(z|S_C) \right)  
\end{aligned}
\end{equation}
where $\mathbb{E}$ denotes the mathematical expectation; $D_{KL}$ represents Kullback-Leibler Divergence; $q(z|S_T)$ and $q(z|S_C)$ represent the variational posterior $q(z|H_T, Y_T, H_C, Y_C)$ and $q(z|H_C, Y_C)$, respectively.

\subsection{Graph Attention Networks (GAT)}
\subsubsection{Graph construction}
In reality, autonomous vehicles are not isolated islands and they need to constantly interact with other road users. As human drivers, they plan their routes by continuously paying attention to the vehicles next to them while driving. Therefore, we construct an undirected graph $G = \{V, E\}$ to capture the relationship between traffic participants.

In this work, we focus on the ego vehicle and surrounding vehicles. Hence, in the graph $G$, a node $v_i$ in node set $V$ represents a vehicle in the highway scenario. The node set $V$ is formulated as: $V = \{v_0, v_1, \ldots, v_n\}$, where $n$ is the number of vehicles in the scenario.

During the data pre-processing phase, we only recorded the vehicles inside a rectangular grid with ego vehicles as the center point, which is 200 feet long and 35 feet wide. Thus, we consider only the vehicles inside the grid when defining the edge set $E$.
Here, we utilize the adjacency matrix $A$ to represent the edges. We define $A_{ij}$ as the weight calculated by the Radial Basis Function (RBF) \cite{buhmann2000radial} if both the $i$-th and $j$-th subgrids contain vehicles; otherwise, $A_{ij}$ is set to 0:
\begin{align}
A_{ij} = \begin{cases} 
\exp\left(-\frac{{\text{dist}(v_i, v_j)}^2}{\delta^2}\right) & \text{if $v_i$, $v_j$ in grid} \\
0 & \text{otherwise.} 
\end{cases}
\end{align}
where $\text{dist}(v_i, v_j)$ represents the straight-line distance between $v_i$ and $v_j$, $\delta$ take the maximum distance in the grid to the center of the grid.

\subsubsection{GAT Structure}
The GAT utilizes graph attentional layers and particular attentional setups to update node features. Initially, a weight matrix $\mathbf{W}$  is applied to every node, followed by the computation of the self-attention mechanism. Then, we use the softmax function to normalize the coefficients, followed by the applications of the LeakyReLU nonlinearity. Through these computational processes, the new coefficients $\alpha_{ij}$ can be expressed as:
\begin{align}
    \alpha_{ij} = \frac{\exp(\text{LeakyReLU}(\mathbf{a}^T[\mathbf{W}S_i \| \mathbf{W}S_j]))}{\sum_{k \in \mathbb{N}_i} \exp(\text{LeakyReLU}(\mathbf{a}^T[\mathbf{W}S_i \| \mathbf{W}S_k]))}
\end{align}
where $\mathbf{a}^T$ represents a weight vector, $(\cdot)^T$ denotes transposition, $\mathbb{N}_i$ denotes the set of nodes associated with node $i$ and $\|$ means the concatenation operation. Finally, we aggregate adjacent node representations with multi-head attention to update each node representation iteratively.

\begin{align}
    {S}'_i = f \left( \frac{1}{K} \sum_{k=1}^{K} \sum_{j \in \mathbb{N}_i} \alpha_{ij}^k W^k {S}_j \right)
\end{align}
where $f$ represents a nonlinear activation function, $K$ means the number of heads of the multi-attention mechanism.

\section{EXPERIMENT}
In this section, we evaluate the performance of our model on highD dataset. Firstly, we provide a brief introduction to the dataset and describe the pre-treatment process. Then, we compare GRANP with the various baseline methods with the Root Mean Squared Error (RMSE) and Negative Log-Likelihood (NLL). Finally, a case study is carried out to demonstrate the rationality of the predicted results by visualizing the predictions we generated with the ground truth.

\subsection{Datasets}
The highD dataset \cite{highDdataset} was collected and processed by a team from Institute for Automotive Engineering using drones. At six different locations, they gathered naturalistic vehicle trajectories and recorded data from more than 110,500 vehicles. In this work, we followed the data processing steps in \cite{song2020pip}.

For an equitable comparison, we forecast trajectories for the next 5 seconds by relying on data from the preceding three seconds, and downsample the trajectories to 5Hz. Additionally, we use the z-score method for normalization.

\subsection{Metrics and Settings}
Our model directly learns parameters of Gaussian distributions, hence can efficiently quantify uncertainty. In this work, we compute the Root Mean Square Error (RMSE) between predicted trajectories and the ground truth using the predicted mean of this distribution. Additionally, for the purpose of quantifying GRANP's ability to accurately predict uncertainty, we also use NLL as an evaluation metric.

The training process is carried out on a system running Ubuntu 20.04, equipped with an A800-SXM4-80GB GPU for computational tasks. The hyperparameters are set to a learning rate of $5\times10^{-4}$ and epochs of 2000, and Adam \cite{kingma2014adam} is used as the optimizer and the ReLU function works as the activation function.

\subsection{Baseline Methods}
To verify the effectiveness of our model, we compare GRANP with several classic and state-of-the-art methods:
\begin{itemize}
    \item S-LSTM \cite{Alahi_2016_CVPR}: Social LSTM applies a Social pooling layer that enables information sharing between individual LSTM models, each representing a separate trajectory.

    \item CS-LSTM \cite{Deo_2018_CVPR_Workshops}: Convolutional Social LSTM replaces the fully connected layer of the Social pooling layer in Social LSTM \cite{Alahi_2016_CVPR} with a convolutional layer.

    \item S-GAN \cite{Gupta_2018_CVPR}: Social GAN uses a Generative Adversarial Network (GAN)-based encoder-decoder framework to generate socially acceptable trajectories.

    \item NLS-LSTM \cite{messaoud2019non}: NLS-LSTM utilizes a non-local multi-head attention mechanism to model the interaction among all the adjacent vehicles.
                           
    \item PiP \cite{song2020pip}: PiP proposed a pip structure, consisting of a planning coupled module, target fusion model, and maneuver-based decoder, for prediction.
\end{itemize}

\subsection{Results}
As shown in Table \ref{Tab:results}, a comparison is shown between GRANP and other state-of-the-art trajectory prediction models on the highD dataset. In short-term prediction (first 1s), GRANP shows slightly less efficiency than PiP. However, in long-term prediction (last 4s), GRANP significantly outperforms the baseline models. For RMSE, GRANP reduces the prediction error by approximately 50\% compared with the PiP model. For NLL, GRANP outperforms PiP with a 70\% advantage. As RMSE measures the average performance of predictions, NLL considers the variance of data. Our results show that GRANP has a significant superiority compared to other baselines when quantifying uncertainties.

\begin{table}[h!]
\centering
\caption{RMSE and NLL in meters over a 5-second prediction horizon for the models.}
\setlength{\tabcolsep}{1.9pt} 
\begin{tabular}{l c c c c c c c c}
\hline
Metric  & Time & S-LSTM & CS-LSTM & S-GAN & NLS-LSTM & PiP & GRANP\\ 
\hline
RMSE (m)  & 1s & 0.19&0.19&0.30&0.20&\textbf{0.17}& 0.41\\
&2s & 0.57&0.57&0.78&0.57&0.52&\textbf{0.44}\\ 
&3s &1.18 &1.16&1.46&1.14&1.05&\textbf{0.70}\\
&4s & 2.00 &1.96& 2.34&1.90&1.76&\textbf{0.88}\\
&5s &3.20 &2.96&3.41&2.91&2.63&\textbf{1.34}\\
\hline
Metric  & Time & S-LSTM & CS-LSTM & S-GAN & NLS-LSTM & PiP &GRANP\\ \hline
NLL (nats)  & 1s & 0.42&0.37&-&-&\textbf{0.14}&0.34\\
&2s &2.58 &2.43&-&-&2.24&\textbf{0.85}\\ 
&3s &3.93 &3.65&-&-&3.48&\textbf{1.16}\\
&4s &4.87 &4.51&-&-&4.33&\textbf{1.37}\\
&5s &5.57 &5.17&-&-&4.99&\textbf{1.53}\\
\hline
\end{tabular}
\label{Tab:results}
\end{table}

\subsection{Case Study}
\begin{figure*}
    \centering
    \includegraphics[width=1\textwidth]{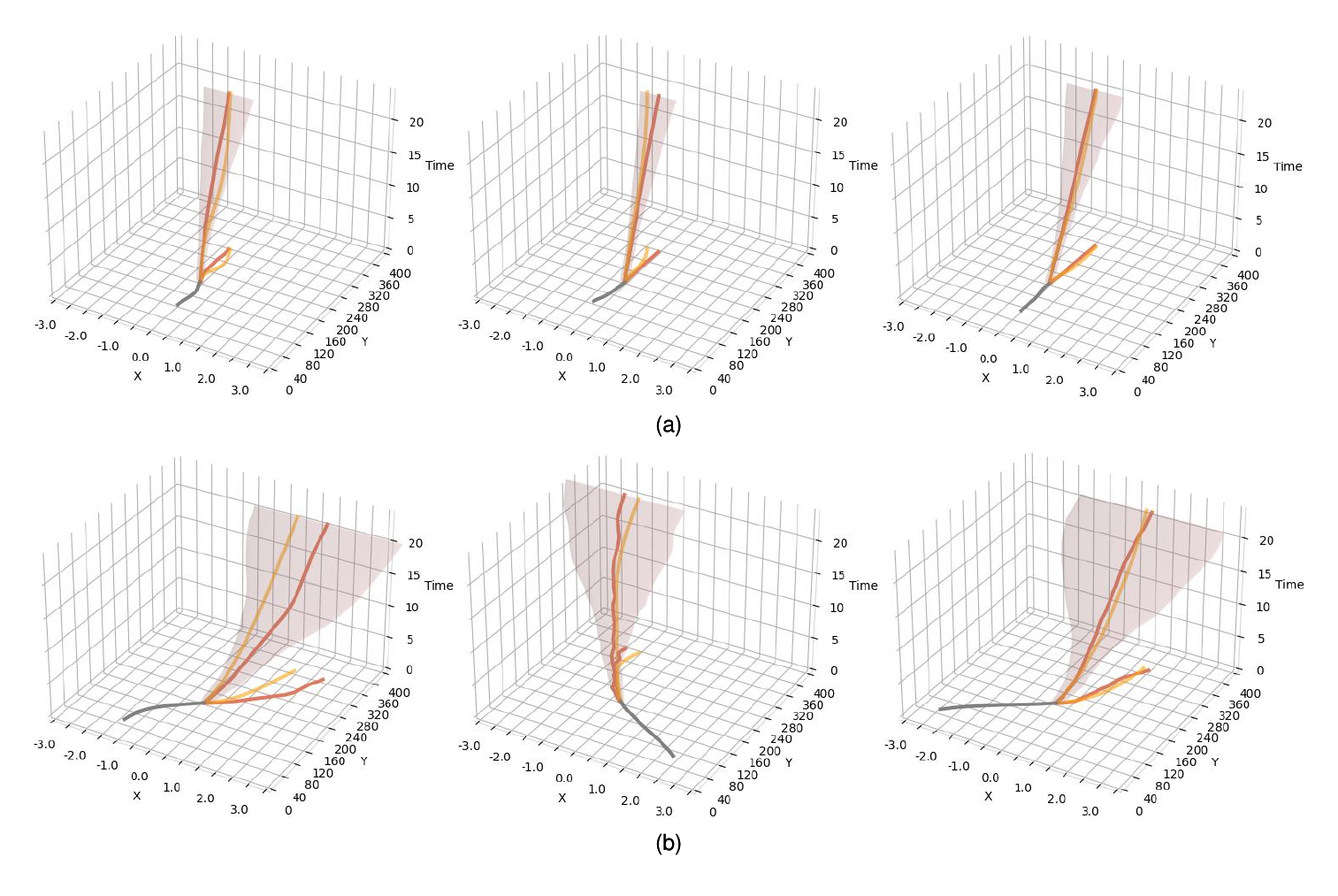}
    \caption{Examples of trajectory prediction of the ego vehicle under (a) going straight and (b) lane-changing scenarios. $x$-axis, $y$-axis, $z$-axis represent the lateral coordinate, longitudinal coordinate, and the time stamp, respectively. The gray, red, and yellow lines represent historical, prediction, and true trajectories, respectively. Additionally, the pink surface represents the uncertainty. In this work, we use 95\% confidence intervals.}
    \label{fig:unt}
\end{figure*}

\begin{figure}
    \centering
    \includegraphics[width=.49\textwidth]{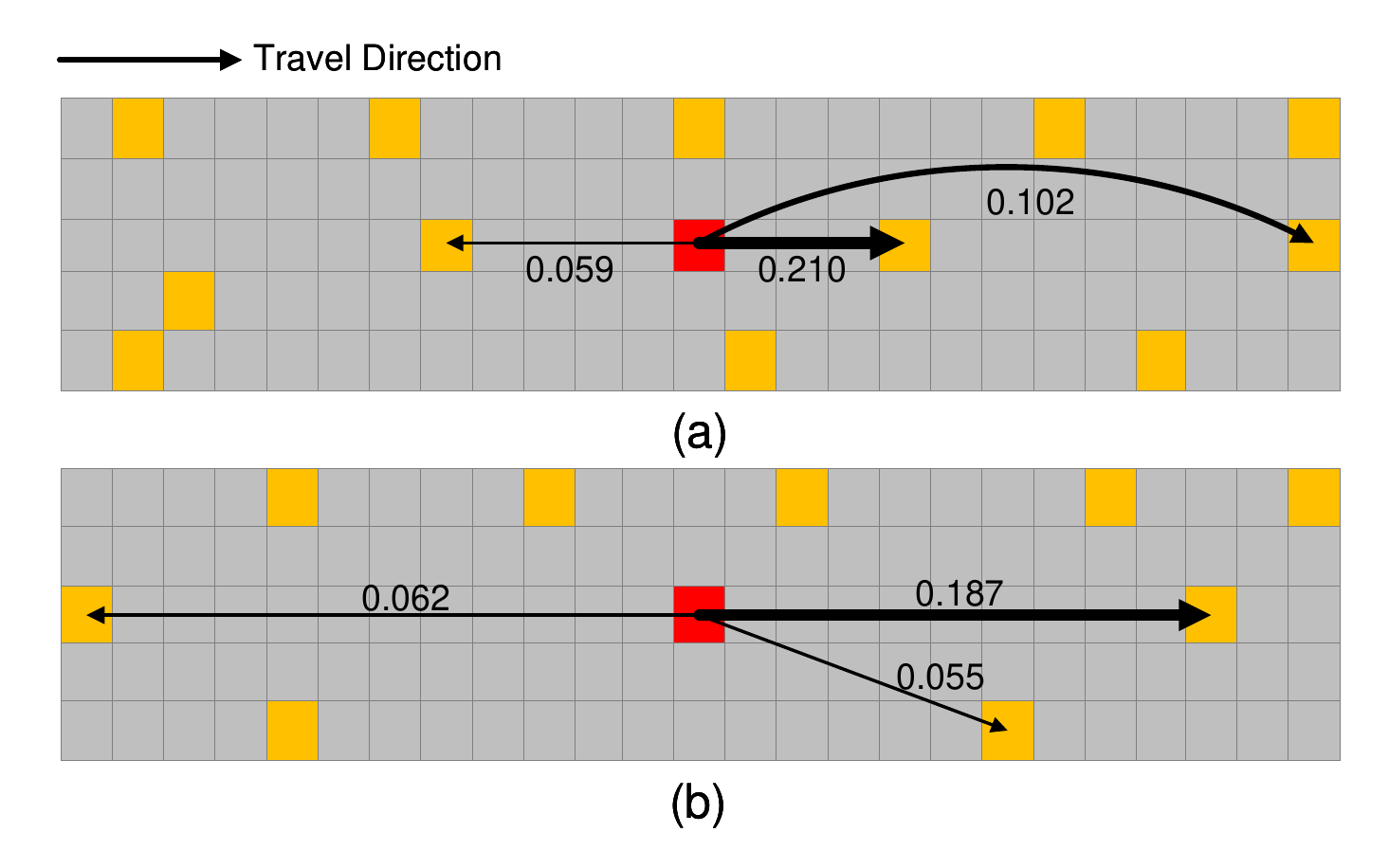}
    \caption{Examples of trajectory prediction of the ego vehicle under (a) going straight and (b) lane-changing scenarios. Red and yellow cells indicate ego and neighboring vehicles respectively. The three most important attention weights are visualized.}
    \label{fig:gat}
\end{figure}

To provide an intuitive understanding that GRANP can predict trajectories accurately and quantify uncertainties in different driving scenarios, we visualize two common driving scenarios, including lane changing and going straight. In Fig. \ref{fig:unt}, the yellow and red lines in the $x$-$y$ plane represent true and predicted trajectories. Obviously, in both lane-changing and straight driving scenarios, the trajectory predicted by GRANP is generally coincident with the real trajectory as shown in Fig. \ref{fig:unt}. 

In Fig. \ref{fig:unt}, the pink surface represents the uncertainty of prediction. For both lane-changing and going straight scenarios, the value of prediction uncertainty increases with time. Additionally, the prediction uncertainty of most lane-changing scenarios is higher than going straight scenarios. The result indicates that lane-changing scenarios are more challenging for trajectory prediction. Some recent studies have considered such issues. For instance, \cite{mersch2021maneuver} utilized a spatial-temporal convolutional network to classify the maneuver intentions first and then predicted trajectories based on intentions. We leave such improvements for future work.

To visualize the interaction among vehicles, we plot the top three attention weights obtained from the GAT. Fig. \ref{fig:gat} (a) and (b) are two cases of going straight and lane-changing respectively. 
In both cases, the ego vehicles assign the highest attention to the leading vehicles, highlighting their significance in the interaction, and the following vehicles also receive attention during the driving process. Moreover, attention is also towards vehicles in the neighboring lane during lane-changing scenarios. The results indicate that GAT can effectively capture social interactions during driving.

\subsection{Sensitivity Analysis}
\begin{figure}[h]
    \centering
    \includegraphics[width=0.49\textwidth]{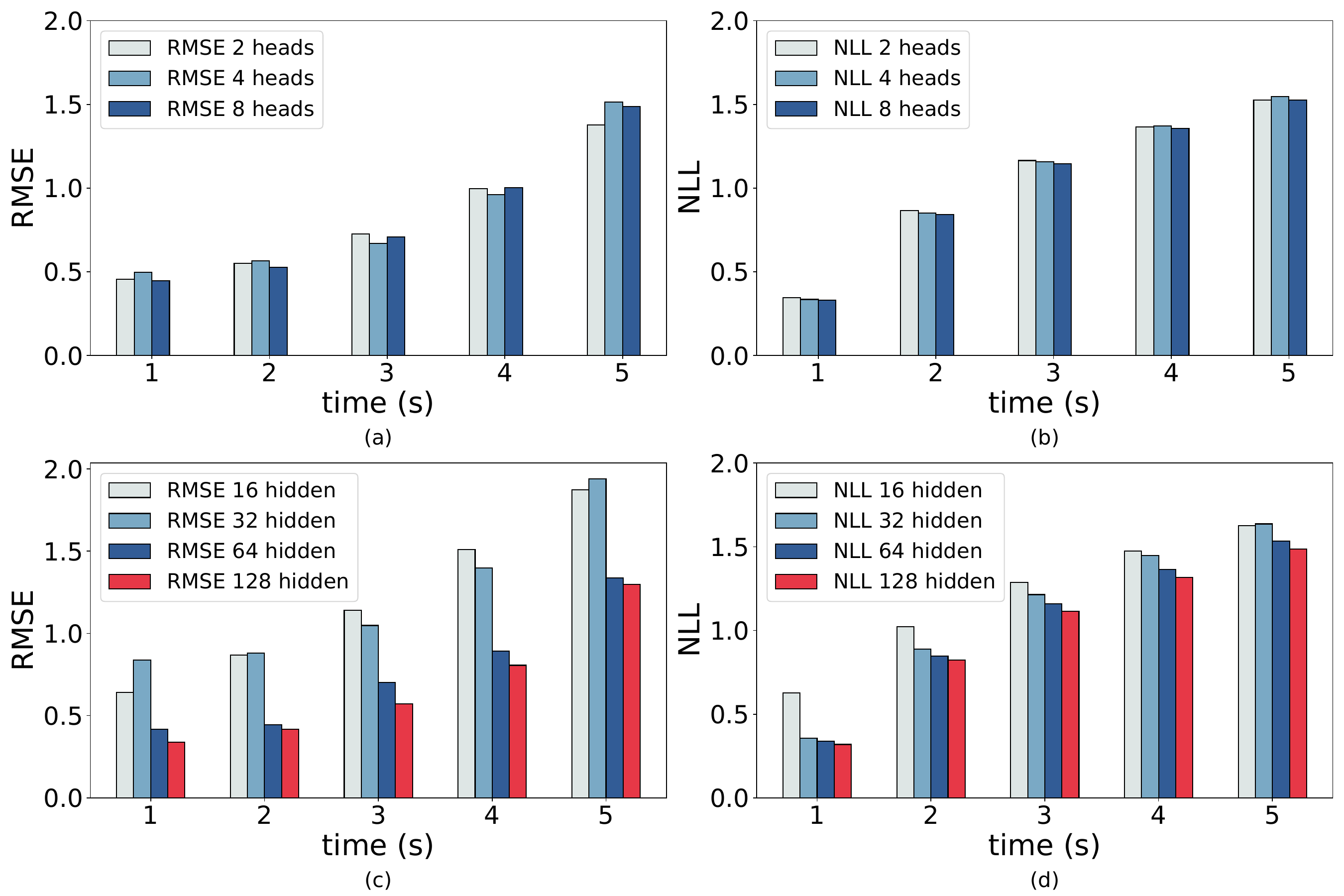}
    \caption{RMSE and NLL results under different parameters. (a) and (b) represents the results with different numbers of attention heads as well as (c) and (d) shows the results with different numbers of hidden dimensions.}
    \label{fig:head}
\end{figure}

In this section, we investigate the impact of various hyperparameters in the model. Specifically, we conduct experiments on hidden dimensions with $\{16, 32, 64, 128\}$, and attention heads with $\{2,4,8\}$. Fig. \ref{fig:head} presents the results. The experiments depicted in Fig. \ref{fig:head} (a) and (b) indicate that our model is not sensitive to attention heads. One possible explanation is that the highD dataset is weakly interactive, and the distances among vehicles are relatively large. 
Therefore, a limited number of attention heads are adequate to capture social interactions.
Regarding the variation of hidden dimensions, the model performance improves with an increase in model complexity. However, we still choose the hidden dimension as 64, prioritizing training efficiency.

\section{CONCLUSIONS}
In this study, we propose a novel model for trajectory prediction named Graph Recurrent Attentive Neural Process (GRANP). Our model includes an encoder with deterministic and latent paths and a decoder for prediction. Compared to previous studies, our method can efficiently quantify uncertainties. The results of extensive experiments on highD show that GRANP significantly outperforms state-of-the-art models. Finally, we present a case study for analysis.

\section*{ACKNOWLEDGMENT}
We would like to acknowledge a grant from the Guangzhou Municipal Science and Technology Project (2023A03J0011); and Guangzhou Basic and Applied Basic Research Project (SL2022A03J00083).

\bibliographystyle{ieeetr}

\bibliography{refs}

\end{document}